\documentclass{sigchi}

\CopyrightYear{2016}
\setcopyright{acmlicensed}
\doi{http://dx.doi.org/10.475/123_4}
\isbn{123-4567-24-567/08/06}
\conferenceinfo{CHI'16,}{May 07--12, 2016, San Jose, CA, USA}
\acmPrice{\$15.00}

\usepackage{balance}       
\usepackage{graphics}      
\usepackage[T1]{fontenc}   
\usepackage{txfonts}
\usepackage{mathptmx}
\usepackage[pdflang={en-US},pdftex]{hyperref}
\usepackage{color}
\usepackage{booktabs}
\usepackage{textcomp}

\usepackage{microtype}        
\usepackage{ccicons}          

\usepackage{todonotes}

\def\plaintitle{Dialogue Design and Management for Multi-Session Casual Conversation with Older Adults}

\def\emptyauthor{}
\def\plainkeywords{Spoken Dialogue Agents; Older Users; Dialogue Management; Casual Conversation; Social Skills Practice.}

\makeatletter
\def\url@leostyle{%
  \@ifundefined{selectfont}{
    \def\UrlFont{\sf}
  }{
    \def\UrlFont{\small\bf\ttfamily}
  }}
\makeatother
\def\pprw{8.5in}
\def\pprh{11in}

\definecolor{linkColor}{RGB}{6,125,233}
\hypersetup{%
  pdftitle={\plaintitle},
  pdfauthor={\emptyauthor},
  pdfkeywords={\plainkeywords},
  pdfdisplaydoctitle=true, 
  bookmarksnumbered,
  pdfstartview={FitH},
  colorlinks,
  citecolor=black,
  filecolor=black,
  linkcolor=black,
  urlcolor=linkColor,
  breaklinks=true,
  hypertexnames=false
}

\begin{document}

\title{\plaintitle}

\numberofauthors{6}
\author{%
  \alignauthor{S. Zahra Razavi\\
    \affaddr{University of Rochester}\\
    \affaddr{Rochester, NY}\\
    \email{srazavi@cs.rochester.edu}}\\
  \alignauthor{Lenhart K. Schubert\\
    \affaddr{University of Rochester}\\
    \affaddr{Rochester, NY}\\
    \email{schubert@cs.rochester.edu}}\\
  \alignauthor{Benjamin Kane\\
    \affaddr{University of Rochester}\\
    \affaddr{Rochester, NY}\\
    \email{bkane2@u.rochester.edu}}\\
  \alignauthor{Mohammad Rafayet Ali\\
    \affaddr{University of Rochester}\\
    \affaddr{Rochester, NY}\\
    \email{mali7@cs.rochester.edu}}\\
  \alignauthor{Kimberly A. Van Orden\\
    \affaddr{University of Rochester}\\
    \affaddr{Rochester, NY}\\
    \email{kimberly\_vanorden@urmc.roches\\ter.edu}}\\
  \alignauthor{Tianyi Ma\\
    \affaddr{University of Rochester}\\
    \affaddr{Rochester, NY}\\
    \email{tma8@u.rochester.edu}}\\
}

\maketitle

\begin{abstract}
 We address the problem of designing a conversational avatar capable of a sequence of casual conversations with older adults. Users at risk of loneliness, social anxiety or a sense of ennui may benefit from practicing such conversations in private, at their convenience. We describe an automatic spoken dialogue manager for LISSA, an on-screen virtual agent that can keep older users involved in conversations over several sessions, each lasting 10-20 minutes. The idea behind LISSA is to improve users' communication skills by providing feedback on their non-verbal behavior at certain points in the course of the conversations. In this paper, we analyze the dialogues collected from the first session between LISSA and each of 8 participants. We examine the quality of the conversations by comparing the transcripts with those collected in a WOZ setting. LISSA's contributions to the conversations were judged by research assistants who rated the extent to which the contributions were ``natural", ``on track", ``encouraging", ``understanding", ``relevant",  and ``polite". The results show that the automatic dialogue manager was able to handle conversation with the users smoothly and naturally.
\end{abstract}

\category{H.5.2.}{Information Interfaces and Presentation
  (e.g. HCI)}{User Interfaces}
  \category{I.2.7.}{Artificial Intelligence}{Natural Language Processing}


\section{Introduction}

 The population of senior adults is growing, in part as a result of advances in healthcare. According to United Nations studies on world population, the number of people aged 60 years and over is predicted to rise from 962 million in 2017 to 2.1 billion in 2050 ~\cite{WorldPopulation2017}. Moreover, large numbers of older adults live alone. According to {\it 2017 Profile of Older Americans}, in the US about 28\% (13.8 million) of noninstitutionalized older persons, and about half of women age 75 and over live alone~\cite{OlderAmericans2017}. Some studies show a significant correlation between depression and loneliness among older people ~\cite{singh2009loneliness}.
 Also many have difficulty managing their personal and household activities on their own and could potentially benefit from technologies that increase their autonomy, or perhaps just to provide engaging interactions in their spare time. 

However, learning to use digital technology can be difficult and frustrating, especially for older people, and this negatively impacts acceptance and use of such technology ~\cite{shaked2017avatars}. But the ever-increasing accuracy of both automatic speech recognition (ASR) systems and text-to-speech (TTS) systems, along with richly engineered apps such as Siri and Alexa, has boosted the popularity of conversational interaction with computers, obviating the need for arcane expertise. Realistic virtual avatars and social robots can make this interaction even more natural and pleasant. 

Spoken language interaction can benefit older people in many different ways, including the following:

\begin{itemize}
    \item Dialogue systems can support people with their health care needs. Such systems can collect and track health information from older individuals and provide comments and advice to improve user wellness. They can also remind people about their medications and doctor appointments.
    \item Speech-based systems can help users feel less lonely by providing casual chat as well as information on news, events, activities, etc. 
    \item They can also provide entertainment such as games, jokes, or music for greater enjoyment of spare time.
    \item Dialogue systems can help older people improve some skills. For instance, cognitive games can help users maintain mental acuity. They can also enable practice of communication skills, as may be desired by seniors experiencing social isolation after loss of connections to close family and friends ~\cite{singh2009loneliness}.
    \item In limited ways, conversational systems can lead ``reminiscence therapy" sessions aimed at reducing stress, depression and boredom in people with dementia and memory disorders, by helping to elicit their memories and achievements ~\cite{arean1993comparative}.
\end{itemize}

All of the above types of systems require a powerful dialogue manager, allowing for smooth and natural conversations with users, and taking account of the special needs and limitations of the target user population.

In this paper we describe how we adapted the dialogue capabilities of a virtual agent, LISSA, for interaction with older adults who may wish to improve their social communication skills. As the centerpiece of the "Aging and Engaging Program" (AEP) ~\cite{ali2018aging}, this version of LISSA holds a casual conversation with the user on various topics likely to engage the user. At the same time LISSA observes and processes the user's nonverbal behaviors, including eye contact, smiling, and head motion, as well as superficial speech cues such as speaking volume, modulation, and an emotional valence derived from the word stream. 

The conversation is in three segments, and at the two break points the system offers some advice to users on the areas they may need to work on and how to improve in those areas. In addition, a summary of weak and strong aspects of the user's behavior is presented at the end of the conversation. 

The system was designed in close collaboration with an expert advisory panel of professionals
(at our affiliated medical research center) 
working with geriatric patients. 
A single-session study was conducted with 25 participants where the virtual agent's contributions to the dialogue were selected by a human wizard. The nonverbal feedback showed an accuracy of 72\% on average, in relation to a human-provided corpus of annotations, treated as ground truth. Also, a user survey showed that users found the program useful and easy to use. 
As the next step, we deployed a fully automatic system where users can interact with the avatar at home. We planned for a 10-session intervention, where in each session the participants have 10-20 minutes of interaction with the avatar. The first and last sessions were held in a lab, where experts collected information on users and rated their communication skills. The remaining sessions were self-initiated by the users at home.  We ran the study with 9 participants interacting with the avatar and 10 participants in a control group. 

Our framework for multi-topic dialogue management was introduced in ~\cite{razavimanaging}. To ensure that the conversations would be appropriate and engaging, our geriatrician collaborators guided us in the selection of topics and specific content contributed to the dialogues by LISSA, based on their experiences in elderly therapy. 
As an initial evaluation of LISSA's conversational behavior we have analyzed 8 in-lab conversations using the automated version of LISSA, comparing these with 8 such conversations where LISSA outputs were selected by a human wizard. In all cases the human participants were first-time users (with no overlap between the two groups). The transcripts were deidentified, randomized, and distributed to 6 research assistants (RAs), with each transcript being rated by at least 3 RAs. Ratings for each transcript were then averaged across the RAs. The results showed that the interactions with the automatic system earned high ratings, indeed on some criteria slightly better than WOZ-based interactions. 

We have initiated further data analyses, and three aspects relating to conversation quality are the following. First, we will be looking at the level of verbosity for different users in different sessions, to determine its dependence on the user's personality and on the topic under discussion. Second, we will measure self disclosure and study its correlation with user mood and personality. Third, we will track users' sentiment over the course of conversations to determine its variability over time, and its dependence on dialogue topics and user personality.

The main contributions of the work reported here are
\begin{itemize}
    \item demonstration of the flexibility of our approach to dialogue design and management, as evidenced by rapid adaptation and scaling up of previous versions of LISSA's repertoire, suitable for the Aging and Engaging domain; our approach uses modifiable dialogue schemas, hierarchical pattern transductions, and context-independent ``gist clause" interpretations of utterances;
    \item integration of an automated dialogue system for multi-topic conversations into a virtual human (LISSA), designed for conversation practice and capable of observing and providing feedback to the user;
    \item an initial demonstration showing that the automated dialogue manager functions as effectively for a sampling of users as the prior wizard-guided version; this signifies an advance of the state of the art in building conversational practice systems for older users, with no constraints on users' verbal inputs (apart from the conversational context determined by LISSA's questions).
\end{itemize}

The rest of this paper is organized as follows. We first comment on extant work on spoken dialogue systems that are designed to help older adults; then we introduce our virtual agent, LISSA, and briefly explain the feedback system. We then discuss the dialogue manager structure and the content design for multi-session interactions with older users. In the next section, we evaluate LISSA's first-session interactions with users, referred to above. We then mention ongoing and future work, and finally summarize our results and state our conclusions.

\section{Related Work}
As noted in the the introduction, thanks in part to recent improvements in ASR and TTS systems, the use of conversational systems to help technically unskilled older adults has become more feasible. 


According to ~\cite{yaghoubzadeh2013virtual}, employing virtual agents as daily assistants for older adults puts at least two questions in front of us: first, whether potential users would accept these systems and second, how the systems can interact with users robustly, while taking into account the limitations and needs of this user population. ~\cite{guide2011} showed that older people who are unfamiliar with technology prefer to interact with assistive systems in natural language. The most important features for effective interaction of virtual agents with a user are discussed in ~\cite{shaked2017avatars}. Key among them is ease of use, especially for older individuals with reduced cognitive abilities. Virtual agents also should have a likable appearance and persona. Moreover, the system should be able to personalize its behavior according to the specific needs and preferences of each user. Another important yet challenging feature is the ability to recover from mistakes, misunderstandings, and other kinds of errors, and subsequently resume normal interaction. 

~\cite{vardoulakis2012designing} found high acceptance of a companionable virtual agent -- albeit WOZ-controlled -- among users, when they could talk about a topic of interest to them. 
Some systems provide a degree of social companionship with older users through inquiries about subjects of interest and providing local information relevant to those interests. For instance ~\cite{miehle2019social} tried single-session interactions where a robot reads out newspapers according to users' topics of interest and asks them some personal questions about their past, attending to the response by nodding and maintaining eye contact. The survey results showed that the participants liked the interaction and were open to further sessions with the system. ~\cite{abdollahi2017pilot} introduced ``Ryan" as a companion who interprets user's emotions through verbal and nonverbal cues. Six individuals who lived alone enjoyed interacting with Ryan and they felt happy when the robot was keeping them company. However, they did not find talking to the robot to be like talking to a person. 
Another application is to gather health information (such as blood pressure and exercise regimen) during the interaction and provide health advice accordingly ~\cite{ono2018virtual}. Unlike the work reported here, none of these projects make an attempt to engage users in topically broad, casual dialogue, to understand users' inputs in such dialogues, or to allow for conversational skills practice, with feedback by the system. %

Virtual companions have also been proposed for palliative care -- helping people with terminal illnesses reduce stress by steering them towards topics such as spirituality, or their life story. In an exploratory study with older adults reported in ~\cite{utami2017talk}, 44 users interacted with a virtual agent about death-related topics such as last will, funeral preparation and spirituality. They used a tablet displaying an avatar that used speech and also some nonverbal signals such as posture and hand gestures; users selected their input utterances from a multiple-choice menu. The study showed that people were satisfied with their interaction, found the avatar understanding and easy to interact with, and were prepared to continue their conversation with the avatar. Again, however, systems of this type so far do not actually attempt to extract meaning from miscellaneous user inputs, apart from answers to inquiries about specific items of information. Nor do they generally provide feedback based on observing the user, or provide opportunities for practicing conversational skills.

Other systems have focused on assisting older adults with their specific daily needs. For instance ~\cite{tsiourti2016virtual} introduces a virtual companion, ``Mary", for older adults that assists them in organizing their daily tasks by responding to their needs such as offering reminders, guidance with household activities, and locating objects using the 3D camera. A 12-week interaction between Mary and 20 older adults showed that the companion was accepted very well, although occasionally verbal misunderstandings and errors caused some user frustration. The authors note that high expectations by users may have limited full satisfaction with the system. ``Billie" ~\cite{kopp2018conversational} is another example of a virtual home assistant that helps users organize their daily schedule. Although the task scope was limited, and the first study was not completely automatic, Billie was designed to handle certain challenges often encountered with spoken language systems, especially with older users, such as verbal misunderstandings and topic shifts. The virtual agent also used gestures, facial expressions and head movement for more natural speaking behavior (for instance for emphasis, or to signal uncertainty). The study results showed that users can effectively handle the interaction. 
However, this system does not track the user's non-verbal behavior, and is not capable of casual conversation on various topics. Instead, it focuses on the specific task of managing the user's daily calendar.

Another approach that has proved effective in ameliorating loneliness and depression in older people is reminiscence therapy ~\cite{arean1993comparative}. A pilot study aimed at implementing a conversational agent that collects and organizes memories and stories in an engaging manner is reported in ~\cite{nikitina2018smart}. The authors suggest that a successful companionable agent needs to possess not only a model of conversation and of reminiscence, but also generic knowledge about events, habits, values, relationships, etc., 
to enable it to respond meaningfully to reminiscences. 

Robotic pets are another interesting technology offered for alleviating loneliness in older people; such pets react to user speech and petting by producing sounds and eye and body movements. Studies ~\cite{sung2015robot} show that such interactions can improve users' communication and interaction skills.

In almost all applications of virtual agents, the system needs to somehow motivate users to engage conversationally with it. (Robotic pets are an exception.) An interesting study of people's willingness to disclose themselves in interacting with an open-domain voice-based chatbot is presented in ~\cite{ravichander2018empirical}. The authors discovered that self-disclosure by the chatbot would motivate self-disclosure by users. Moreover, initial self-disclosure (or otherwise) by users is apt to characterize their behavior in the rest of the conversation. For instance, users who self-disclose initially tend to have longer turns throughout the rest of the dialogue, and are more apt to self-disclose in later turns. The authors were unable to confirm a clear link between a tendency towards self-disclosure and positive evaluation of the system. However, they found that people enjoy the conversation more if the agent offers its own backstory. 

\section{The LISSA Virtual Agent}
\begin{figure}
\centering
  \includegraphics[width=0.9\columnwidth]{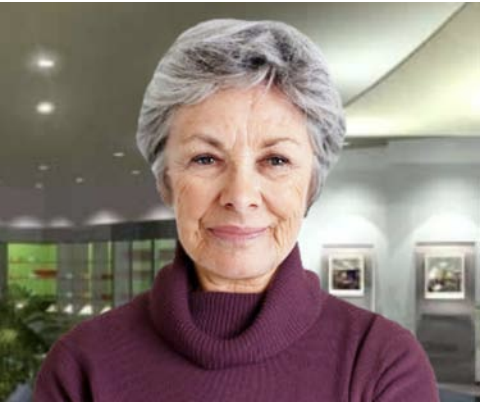}
  \caption{``Aging and Engaging" Virtual Agent}~\label{fig:figure1}
\end{figure}

The LISSA (Live Interactive Social Skills Assistance) virtual agent is a human-like avatar (Figure ~\ref{fig:figure1}) intended to become ubiquitously available for helping people practice their social skills. LISSA can engage users in conversation and provide both real-time and post-session feedback.

\subsection{The system}
To help users improve their communication skills, LISSA provides feedback on their nonverbal behavior, including \textit{eye contact}, \textit{smiling}, \textit{speaking volume}, as well as one verbal feature: the emotional \textit{valence of the word content}. In the original implementation of LISSA, feedback was presented in real time through screen icons that turn from green to red, indicating that the user should improve the corresponding nonverbal behavior. Feedback was also presented at the end of each interaction, via charts and figures providing information on the user's performance. However, as both real-time visual feedback and charts could be cognitively demanding for older users, the Aging and Engaging version of LISSA offers feedback through speech and text during the conversation. Furthermore, at the end of each conversation session, the current system generates a speech- and text-based summary of the feedback provided during the dialogue, mentioning users' strong areas, weaknesses and suggestions for improvement. More details on how feedback is generated and presented can be found in ~\cite{ali2015lissa}.

\subsection{Previous LISSA Studies}
Two versions of LISSA were previously designed based on the needs and limitations of the target users. LISSA showed potential for impacting college students' communication skills when it was tested as a speed-dating partner in a WOZ setting \cite{ali2015lissa}, offering feedback to the user. Subsequently, a fully automatic version was implemented, designed to help high-functioning teens with autism spectrum disorder (ASD) to practice their conversational skills. Experiments were conducted with 8 participants, who were asked to evaluate the system. The ASD teens not only handled the interaction well, but also expressed interest in further sessions with LISSA to improve their social skills ~\cite{razavi2016lissa}. These results encouraged us to improve the system and to prepare for additional experiments with similar subjects. The quest for additional participants, and for evaluation of conversational practice with LISSA as an effective intervention for ASD teens, is still ongoing. At the same time the preliminary successes with LISSA led to the idea of creating a new version with a larger topical repertoire, to be tested in multiple sessions, eventually at home rather than in the lab. The idea was implemented as the ``Aging and Engaging Program". 
\begin{figure}
\centering
  \includegraphics[width=0.95\columnwidth]{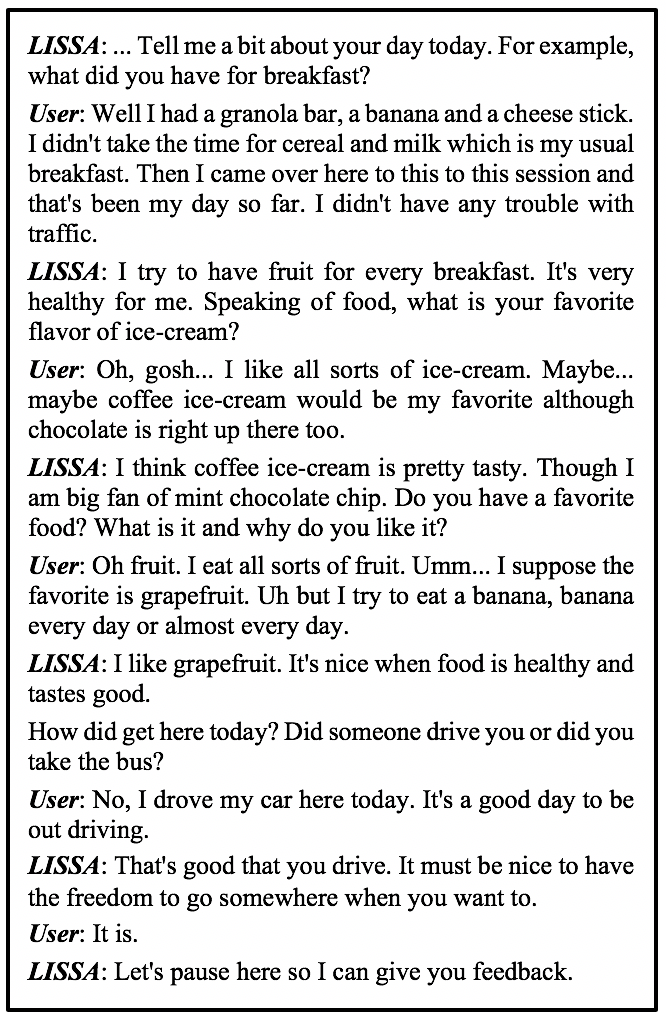}
  \caption{An example of dialogue between LISSA and a user}~\label{fig:figure5}
\end{figure}
\subsection{Aging and Engaging: The WOZ Study}
Social communication deficits among older adults can cause significant problems such as social isolation and depression. Technological interventions are now seen as having potential even for older adults, as computers, cellphones and tablets have become more accessible and easier to use. Our most recent version of LISSA is designed to interact with older users over several sessions, with the goal of helping them improve their communication skills. The system offers low technological complexity and the feedback is presented in a format imposing minimum cognitive load. The interface was designed with the assistance of experts working with geriatric patients, focusing on 12 older adults. 

The program is designed so that each user becomes engaged in casual talk with the system, where two or three times during the conversation, the system suspends the conversation and comments on the user's weaknesses and strengths, and suggests ways of improving on the weaker aspects. Also at the end of the conversation, the system briefs the user on what they went through and summarizes the previously offered advice.

To evaluate the program's potential effectiveness, we conducted a one-shot study with 25 older adults (all more than 60 years old, average age 67, 25\% male), using a WOZ setting for handling the conversation. The results showed that the participants' speaking times in response to questions, as well as the amount of positive feedback, increased gradually in the course of conversation. At the same time, participants found the feedback useful, easy to interpret, and fairly accurate, and expressed their interest in using the system at home. More details on the study can be found in ~\cite{ali2018aging}. 

\subsection{Multi-session Automatic Interaction with Older Adults}
Based on the outcome of the WOZ studies, we designed a multi-session study where participants can interact with the system in their homes at their convenience. The purpose is to study the acceptability of a longer-term interaction and its impact on users' communication skills. The design enables each participant to engage in ten conversation sessions with the avatar. The first and the last sessions are held in the lab, where users fill out surveys and are evaluated for their communication skills by experts. The rest of the sessions are held in users' homes where they need to have access to a laptop or a personal computer.

Each interaction with LISSA consists of three subsessions where LISSA leads a natural casual conversation with the user. Users receive brief feedback on their behavior, during two breaks between subsessions. At the end, users are provided with a behavioral summary, and some suggestions on how they can improve their communication skills. The study has been under way with 9 participants using the avatar to improve their skills. We also recruited 10 participants as a control group. The data has been collected and partially analyzed. 

In the next section, we provide some details on how the system manages an engaging conversation with the user, along with an evaluation of the quality of the conversations gathered. Figure ~\ref{fig:figure5} shows a portion of a conversation between a user and LISSA in the first session of interaction.

\section{The LISSA Dialogue Manager} 
In order to handle a smooth natural conversation on everyday topics with a user, the dialogue manager (DM) needs to follow a coherent plan around a topic. It needs to extract essential information even from relatively wordy inputs, and produce relevant comments demonstrating its understanding of the user's turns. The DM also needs a way to respond to user questions, and to update its plan if this is necessitated by a user input; for instance, if a planned query to the user has already been answered by some part of a user's input, that query should be skipped. The LISSA DM is capable of such plan edits, as well as expansion of steps into subplans. We now explain the DM structure along with the content we provided for the Aging and Engaging program.

\subsection{The DM Structure}
Our approach to dialogue management is based on the hypothesis that human cognition and
behavior rely to a great extent on dynamically modifiable schemas ~\cite{schank2013scripts}~\cite{van1983strategies} and on hierarchical pattern recognition/transduction. 

Accordingly the dialogue manager we have designed follows a flexible, modifiable dialogue schema, specifying a sequence of intended and expected interactions with the user, subject to change as the interaction proceeds. The sequence of formal assertions in the body of the schema express either actions intended by the agent, or inputs expected from the user. These events are instantiated in the course of the conversation. This is a simple matter for explicitly specified utterances by the agent, but the expected inputs from the user are usually specified very abstractly, and become particularized as a result of input interpretation. Schemas can be extended allow for specification of participant types, preconditions, concurrent conditions, partial action/event ordering, conditionals, and iteration. These more general features are still in the early stages of implementation, but so far have not been needed for the present application. 

Based on the main conversational schema, LISSA leads the main flow of the conversation by asking the user various questions. Following each user's response, LISSA might show one of the following behaviors:
\begin{itemize}
    \item Making a relevant comment on the user's input;
    \item Responding to a question, if the user asked one (typically, at the end of the input);
    \item Instantiating a subdialogue, if the user switched to an ``off-track" topic (an unexpected question may have this effect as well.)
\end{itemize}

A dialogue segment between LISSA and one user in the Aging program can be seen in Figure ~\ref{fig:figure5}.

\begin{figure}
\centering
  \includegraphics[width=0.9\columnwidth]{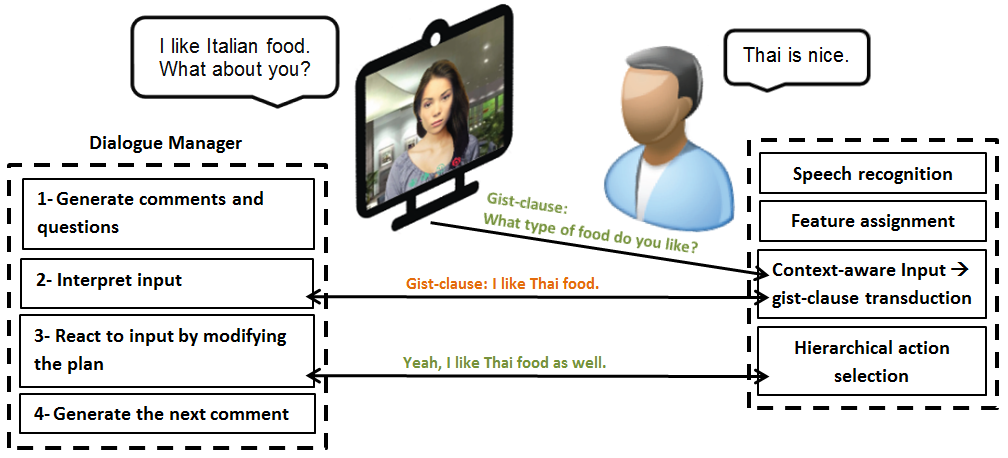}
  \caption{Overview of the dialogue management method}~\label{fig:figure2}
\end{figure}

Throughout the conversation, the replies of the user are transduced into simple, explicit, largely context-independent English sentences. We call these \textit{gist-clauses}, and they provide much of the power of our approach. The DM interprets each user's input in the context of LISSA's previous question, using this question to select pattern transduction hierarchies relevant to interpreting the user's response.  It then applies the rules in the selected hierarchies to derive one or more gist-clauses from the user's input. The transduction trees are designed so that we extract as much information as we can from a user input. The terminal nodes in the transduction trees are templates that can output gist-clauses. As an example of a gist clause, if LISSA asks "Have you seen the new Star Wars movie?" and the user answers "Yes, I have.", the output of the choice tree for interpreting this reply would be something like "I have seen the new Star Wars movie."

The system then applies hierarchical pattern transduction methods to the gist-clauses derived from the user's input to generate a specific verbal reaction. Figure ~\ref{fig:figure2} shows an overview of the dialogue manager. The most important stratagem in our approach is the use of questions asked by LISSA as context for transducing a gist-clause interpretation of the user's answer, and in turn, the use of those gist-clause interpretations as context for transducing a meaningful reaction by LISSA to the user's answer (or multiple reactions, for instance if the user concludes an answer with a reciprocal question). The interesting point is that in gist-clause form, question-answer-reaction triples are virtually independent of the conversational context, and this leads to ``portability" of the infrastructure for many such triples from one kind of (casual) dialogue to another. More details can be found in ~\cite{razavimanaging}.

\subsection{The Content}
To enable a smooth, meaningful conversation with the user, the pattern transduction trees need to be designed carefully and equipped with appropriate patterns at the nodes. For the Aging and Engaging program we needed ten interactions between LISSA and each user. Each interaction consists of three subsessions, and each subsession contains 3-5 questions to the user, along with sporadic self-disclosures by LISSA. In these disclosures LISSA presents herself as a 65-year old widow who moved to the city a few years ago to live with her daughter, and relates relevant information or experiences of hers. LISSA leads the conversation by opening topics, stating some personal thoughts and encouraging user inputs by asking questions. Upon receiving such inputs, LISSA makes relevant comments and expresses her thoughts and emotions.   

LISSA's contributions to the dialogues, including the choice of the character and her background, were meticulously designed by 4 trained research assistants (RAs), in extensive consultation with gerontologists with expertise in interventions. Details of her character, interests, beliefs and thoughts were designed and inserted into the interaction along the way. The contents of the DM's transduction trees were designed based on the suggestions provided by our expert collaborators, as well as on the experience gathered from previous LISSA-user interactions in the WOZ study.

\begin{figure}
\centering
  \includegraphics[width=0.9\columnwidth]{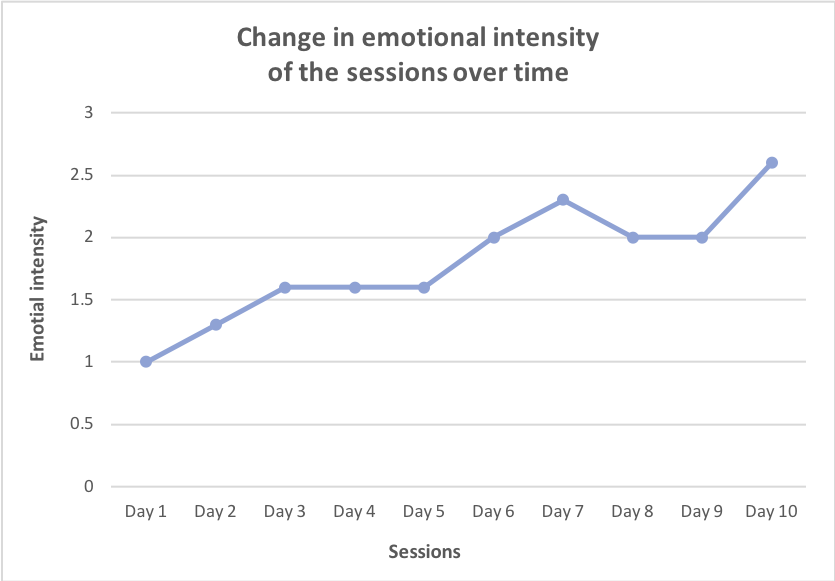}
  \caption{Average emotional intensity of interactions based on the topics of conversation. (Easy = 1, Medium = 2, Hard = 3)}~\label{fig:figure3}
\end{figure}

Since we planned for ten interactions between LISSA and each user, we collected a list of 30 topics that could be of interest to our target group. The gerontological experts divided the topics into three groups based on their emotional intensity or degree of intimacy: easy, medium, and hard. All the topics were ones that older people would encounter in their daily lives. While the ``easy" ones involve little personal disclosure or intimacy and are typical of conversations they might have at a senior center with new people or at a dining hall in a senior living community, the harder conversations contain more emotionally evocative topics.  Table ~\ref{tab:table1} shows a list of topics in each group. We designed each conversation for smooth topic transitions from easier topics in the first session to progressively more challenging topics in the later sessions. Figure ~\ref{fig:figure3} shows how the emotional intensity increases as we move through the study.

\begin{table*}[t]
  \centering
  \begin{tabular}{l l}
     \multicolumn{2}{c}{} \\
    {\textit{Emotional intensity}}   & {\small \textit{Topics}} \\
   \hline
    Easy & Getting to know each other, Where are you from?, City where you live (I), Activities and hobbies, Pets, \\ 
     & Family and friends, Weather, Plan for rest of the day, Household chores \\
    Medium & Holidays and gathering, Cooking, Travel, Outdoor activities, City where you live (I), Managing money, \\ 
     & Education,  Employment and retirement, Books and Newspapers, Dealing with technology, The arts,  \\ 
     & Exercise, Health, Sleep problems, Home \\
    Hard & Tell me about yourself (hopes, wishes, qualities), Driving, Life goals, Growing older, Staying active, \\ 
     & Spirituality 
   
  \end{tabular}
  \caption{Conversation topics and their emotional intensity in Aging and Engaging Program}~\label{tab:table1}
\end{table*}

It is noteworthy that the design and implementation of the infrastructure for the 30 topical dialogues represented an order-of-magnitude scale-up from the previous dialogue designs for speed-dating and for interaction with autistic teens. This testifies to the effectiveness of our use of schemas, transduction hierarchies, and interpretations in the form of gist clauses for rapid dialogue implementation. The individual topics were mostly implemented by three of the authors, including two undergraduates, and typically took about half a day or a day to complete. 

\section{Conversation evaluation}

To evaluate the efficacy of the DM system, we compared conversation transcripts from the initial WOZ-based study with transcripts from the fully automatic dialogue system. To do this, 16 conversation transcripts were chosen, each consisting of a single session with three subtopics. 8 transcripts (the first 8 that became available) were pulled from the WOZ study, and the other 8 were pulled from the automated sessions. These transcripts were de-identified, assigned random numerical labels, and cleaned so that the transcripts followed a uniform format.

The transcripts were then assigned randomly to 6 RAs, with each RA being responsible for assessing 3 transcripts. Each RA was tasked to rate each assigned transcript on six criteria related to the quality of the conversation, which are shown in Table ~\ref{tab:table2}. Each criterion was independently rated on a scale from 1 (not at all) to 5 (completely). No other guidance was provided to the RAs; rather they were directed to make natural judgments for each criterion based on the transcript alone.

\begin{figure}
\centering
  \includegraphics[width=0.9\columnwidth]{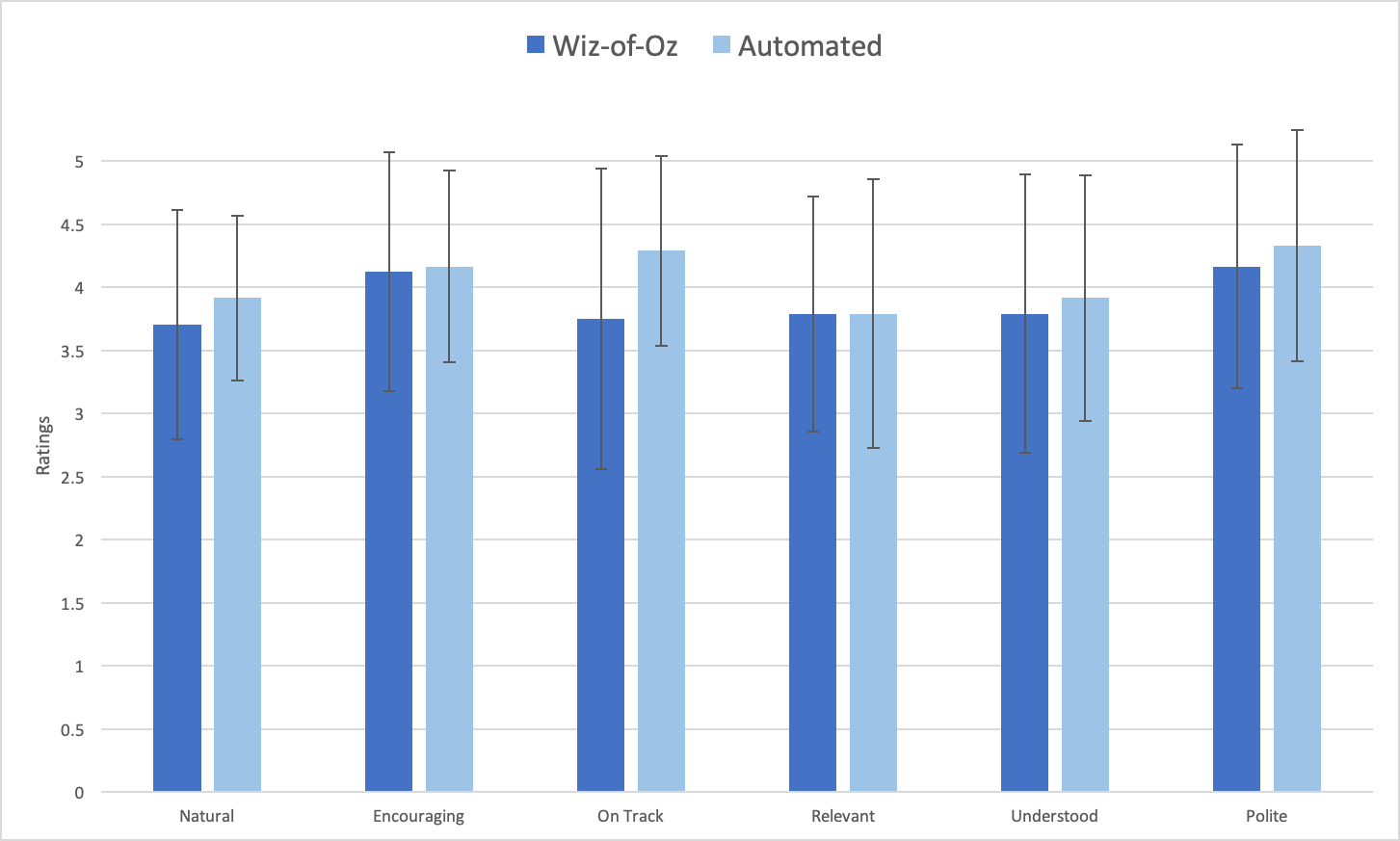}
  \caption{Means and Standard Deviations of Dialogue Ratings for WOZ vs. Automated System}~\label{fig:figure4}
\end{figure}

Ratings across RAs were then averaged for each transcript to represent the consensus score for that transcript. The average rating and one standard deviation for the WOZ study compared to the automated LISSA system are shown for each criterion in Figure ~\ref{fig:figure4}.

\begin{table*}[t]
  \centering
  \begin{tabular}{l l l l l}
    & \multicolumn{3}{c}{} \\
    \cmidrule(r){3-4}
    {\small\textit{Criterion}}
    & {\small \textit{WOZ Average}}
    & {\small \textit{WOZ SD}}
    & {\small \textit{LISSA Average}}
    & {\small \textit{LISSA SD}} \\
    \midrule
    How natural were LISSA's contributions to the conversation? & 3.7 & 0.91 & 3.9 & 0.65 \\
    LISSA's questions/comments encourage the user's participation. & 4.1 & 0.95 & 4.2 & 0.76 \\
    The conversation stayed on track. & 3.8 & 1.19 & 4.3 & 0.75 \\
    LISSA's responses were relevant to the conversation. & 3.8 & 0.93 & 3.8 & 1.06 \\
    LISSA understood what user said. & 3.8 & 1.10 & 3.9 & 0.97 \\
    LISSA's responses were polite and respectful. & 4.2 & 0.96 & 4.3 & 0.92 \\
   
  \end{tabular}
  \caption{Average Ratings of Each Criterion for the Automated LISSA System}~\label{tab:table2}
\end{table*}

As can be seen in the evaluation results, the automated LISSA system was able to achieve fairly high ratings for each of the criteria. This suggests that the system was able to hold conversations that were of good overall quality. Compared to the WOZ study, the automated system was able to achieve slightly higher ratings, although none of the gains were sufficiently large to be statistically significant. Nonetheless, these results suggest that the automated LISSA system is capable of producing conversations that are approximately of the same quality as the human-operated (i.e., WOZ) system.

The largest difference between the WOZ ratings and the automated LISSA ratings was in the ``On Track" metric, where the average LISSA rating was half a point higher than the WOZ average. Again, this improvement is not statistically significant, given the variance and low number of samples, but it is a result consistent with the overall design of the system. Since the DM was designed to follow a well-established conversation plan, while also being equipped with some mechanisms to handle off-topic input and then return to the main track, we expected the system to be particularly effective at remaining on track as compared to the human-operated systems.

The good scores we achieved for the ``Encouraging" and ``Polite" metrics strike us as significant. They suggest that the guidance provided by psychiatric experts on the design of the topics, questions, and comments was advantageous. Additionally, our strategy of including self-disclosure and a backstory in the LISSA dialogues was probably a factor in the successful performance of our system. As mentioned in the Related Work section, self-disclosure by an avatar tends to encourage self-disclosure by the users ~\cite{ravichander2018empirical}, and enjoyment is increased if the avatar is provided with a backstory.

\section{Usability of the system}
The results discussed above indicate that LISSA's dialogue system is functioning well. Confirmation of LISSA's potential for improving users' communication skills will require additional study, including: transcription and analysis of the ASR files from the at-home sessions; expert evaluation of users' communication behaviors in the initial and final lab interviews and surveys that were conducted; and additional, longer-range studies with participants in the target population. However, we have been able to obtain some insights into people's feelings about their interaction with LISSA after completing all 10 sessions.

The surveys administered at the end asked participants numerous questions about their assessment of LISSA, including various aspects of the nonverbal feedback they had received throughout the sessions. A subset of questions of interest here concerned the usability of the system. In particular, participants were asked to rate the following statements:
\begin{itemize}
    \item I thought the program was easy to use.
    \item I would imagine that most people would learn to use this program very quickly.
    \item I felt very confident using the program.
    \item Overall, I would rate the user-friendliness of this product as:...
\end{itemize}
Responses were given on a five-point scale ranging from "strongly disagree" (=1) to "strongly agree" (=5).
The results from nine participants showed an average score of 4.33 (sd = 0.67) for the first statement, meaning that they found the system easy to use. The second statement averaged 3.88 (sd = 1.27). The average score on the third statement was 4.25 (sd = 0.43), indicating that participants felt no anxiety in handling the interaction at home on their own; and along similar lines, the ``user friendliness" rating of 4.56 (sd = 0.50) suggests that users had no real difficulties in dealing with the system.

We also provided some qualitative questions about users' opinions about the system and their experience with it. One question asked what they liked about the system. Some dialogue-related responses were the following:\\
- \textit{``having someone to talk to, even though it was a computer"} \\
- \textit{``starting out with a topic that fit in well with lifestyle that allowed responses and conversation, and providing feedback about responses"} \\
- \textit{``I liked her personality and her calmness"} \\
- \textit{``questions were relatable and answerable"} \\
- \textit{``conversational topics were fitting, though at times seemed more elementary"} \\
- \textit{``simple and straightforward"} \\
- \textit{``interfacing with LISSA was like having someone in my home visiting"} \\
- \textit{``the topics seemed general enough and enjoyed talking about them; helped with self-reflection and have deeper conversations rather than just shallow, surface-level conversations"}

When asked what they didn't like about the system, few users registered any complaints about the dialogue per se. One participant wanted longer conversations, commenting: \textit{``a bit more time [should be] devoted to the conversations - ask 4-5 questions (rather than just 3) to help participants feel more comfortable with the conversation"}. One commented that \textit{``having humans instead of computers may be more motivating"} -- a salient reminder that computers remain tools, not surrogate humans.
 

\section{Future work}
We plan to use the conversational data collected from multi-session interactions between LISSA and older users to gain a deeper understanding of certain key aspects of these interactions. These aspects include the following:

\begin{itemize}
    \item \textit{Verbosity}: We plan to study the level of verbosity of participants from session to session, and also the variability of verbosity among participants. Questions of interest are whether verbosity is dependent on the user's personality, the topic and questions asked, and other factors; and also whether verbosity is correlated with a positive view of the system.

    \item \textit{Self-disclosure}: Self-disclosure is considered a sign of conversational engagement and trust. We intend to use known techniques for measuring self-disclosure in a dialogue to assess whether the extent of self-disclosure is more dependent on the user's personality or the inputs from the avatar; and, how self-disclosure might correlate with the user's mood, social skills and evaluation of the system.
    
    \item \textit{Sentiment}: As this program targets older adults who are at risk of isolation and depression, we want to know more about users' moods, based on the emotional valence of their inputs. We plan to track sentiment over the course of successive sessions, taking into account the context of the dialogue (e.g., the topic under discussion). We should then be able to detect any correlations between the users' moods, and their evaluation of the system at various points.

\end{itemize}

\section{Conclusion}
We introduced the design and implementation of a spoken dialogue manager for handling multi-session interactions with older adults. The dialogue manager leads engaging conversations with users on various everyday and personal topics. Users receive feedback on their nonverbal behaviors including eye contact, smiling, and head motion, as well as emotional valence. The feedback is presented at two topical transition points within a conversation as well as at the end. The goal of the system is to help users improve their communication skills. 

The DM was implemented based on a framework proposed in previous work, employing flexible schemas for planned and anticipated events, and hierarchical pattern transductions for deriving ``gist clause" interpretations and responses. We prepared the dialogue manager for 10 interaction sessions, each covering 3 topics. Our DM framework allowed quite rapid development of the 30 topics. The content was adapted to the needs and limitations of older adults, based on collaboration with an expert advisory panel of professionals at our affiliated medical research center. A broad spectrum of topics was established, and the sessions were designed to progress from emotionally undemanding ones to ones calling for greater emotional disclosure.

 We ran a study including 8 participants interacting with the virtual agent. To evaluate the quality of the conversations, we compared the conversations of these participants in the initial in-lab sessions with 8 such conversations where LISSA outputs were selected by a human wizard. The transcripts were randomly assigned to 6 RAs, who rated the transcripts based on 6 features of high-quality conversation. The results showed high ratings for the automatic system, even slightly better than wizard-moderated interactions. The overall usability evaluation of the system by users who completed the series of sessions shows that users found the system easy to use and rated it as definitely user-friendly. Further studies of interaction quality and effects on users, based (among other data) on ASR transcripts of the at-home sessions, are in progress.

 
\balance{}

\balance{}

\bibliographystyle{SIGCHI-Reference-Format}


\end{document}